\documentclass[final]{cvpr}

\usepackage{times}
\usepackage{epsfig}
\usepackage{graphicx}
\usepackage{amsmath}
\usepackage{amssymb}
\usepackage{xcolor}
\usepackage{multirow}
\usepackage{enumitem}
\usepackage{float}
\usepackage{blindtext}
\usepackage{caption}

\usepackage[pagebackref=true,breaklinks=true,colorlinks,bookmarks=false]{hyperref}

\def\cvprPaperID{4905} 
\def\confYear{CVPR 2021}

\begin{document}

\title{LiveView: Dynamic Target-Centered MPI for View Synthesis}

\author{Sushobhan Ghosh\\
Northwestern University\\
{\tt\small sushobhanghosh@u.northwestern.edu}
\and
Zhaoyang Lv\\
Facebook Reality Labs\\
{\tt\small zhaoyang@fb.com}
\and
Nathan Matsuda\\
Facebook Reality Labs\\
{\tt\small nathan.matsuda@fb.com}
\and
Lei Xiao\\
Facebook Reality Labs\\
{\tt\small lei.xiao@fb.com}
\and
Andrew Berkovich\\
Facebook Reality Labs\\
{\tt\small andrew.berkovich@fb.com}
\and
Oliver Cossairt\\
Northwestern University\\
{\tt\small oliver.cossairt@northwestern.edu}
}

\makeatletter
\let\@oldmaketitle\@maketitle
\renewcommand{\@maketitle}{\@oldmaketitle
    \includegraphics[width=\linewidth, keepaspectratio]{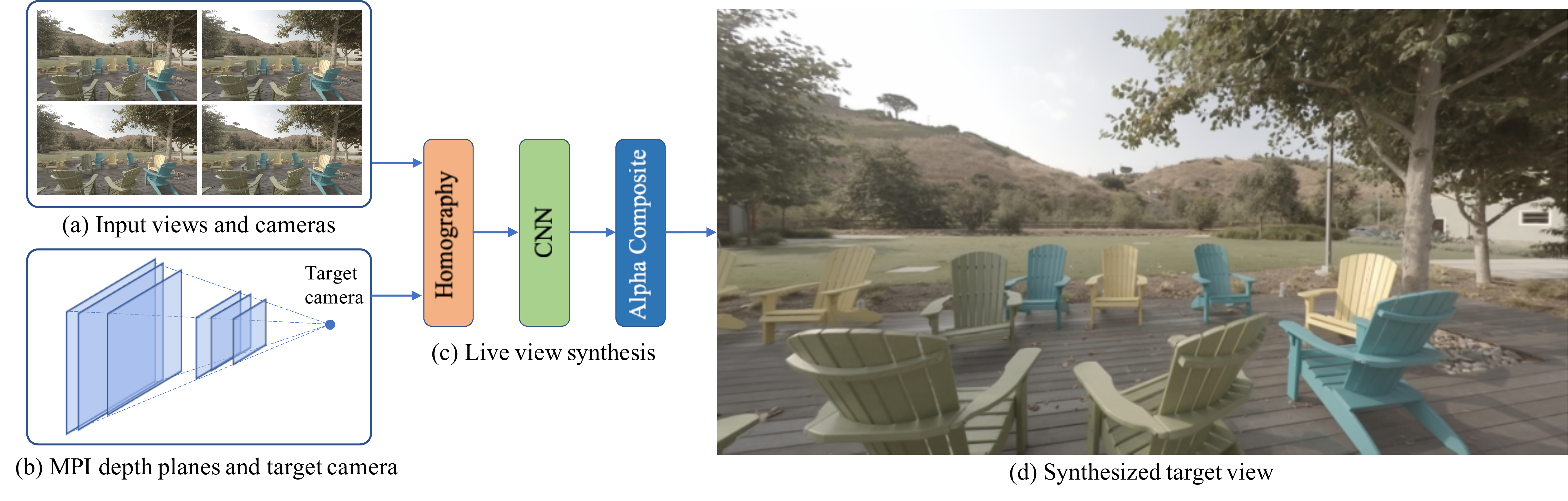}
    \captionof{figure}{\textbf{An overview of our method}. A set of input images with known camera positions (a) and target camera and MPI plane locations (b) are passed to our live view synthesis network (c), which processes the input to produce a synthesized target view (d).}
    \label{fig:intro}
    \vspace{3mm}
}

\makeatletter
\renewcommand{\paragraph}{%
  \@startsection{paragraph}{4}%
  {\z@}{1ex}{-1em}%
  {\normalfont\normalsize\bfseries}%
}
\makeatother

\maketitle
\begin{abstract}
Existing Multi-Plane Image (MPI) based view-synthesis methods generate an MPI aligned with the input view using a fixed number of planes in one forward pass.
These methods produce fast, high-quality rendering of novel views, but rely on slow and computationally expensive MPI generation methods unsuitable for real-time applications.
In addition, most MPI techniques use fixed depth/disparity planes which cannot be modified once the training is complete, hence offering very little flexibility at run-time.

We propose LiveView - a novel MPI generation and rendering technique that produces high-quality view synthesis in real-time. Our method can also offer the flexibility to select scene-dependent MPI planes (number of planes and spacing between them) at run-time.
LiveView first warps input images to target view (target-centered) and then learns to generate a target view centered MPI, one depth plane at a time (dynamically). 
The method generates high-quality renderings, while also enabling fast MPI generation and novel view synthesis.
As a result, LiveView enables real-time view synthesis applications where an MPI needs to be updated frequently based on a video stream of input views.
We demonstrate that LiveView improves the quality of view synthesis while being 70$\times$ faster at run-time compared to state-of-the-art MPI-based methods. 
\end{abstract}


\vspace{-3mm}
\section{Introduction}
\label{sec:introduction}

\begin{figure}
    \centering
    \includegraphics[width=0.45\textwidth, keepaspectratio]{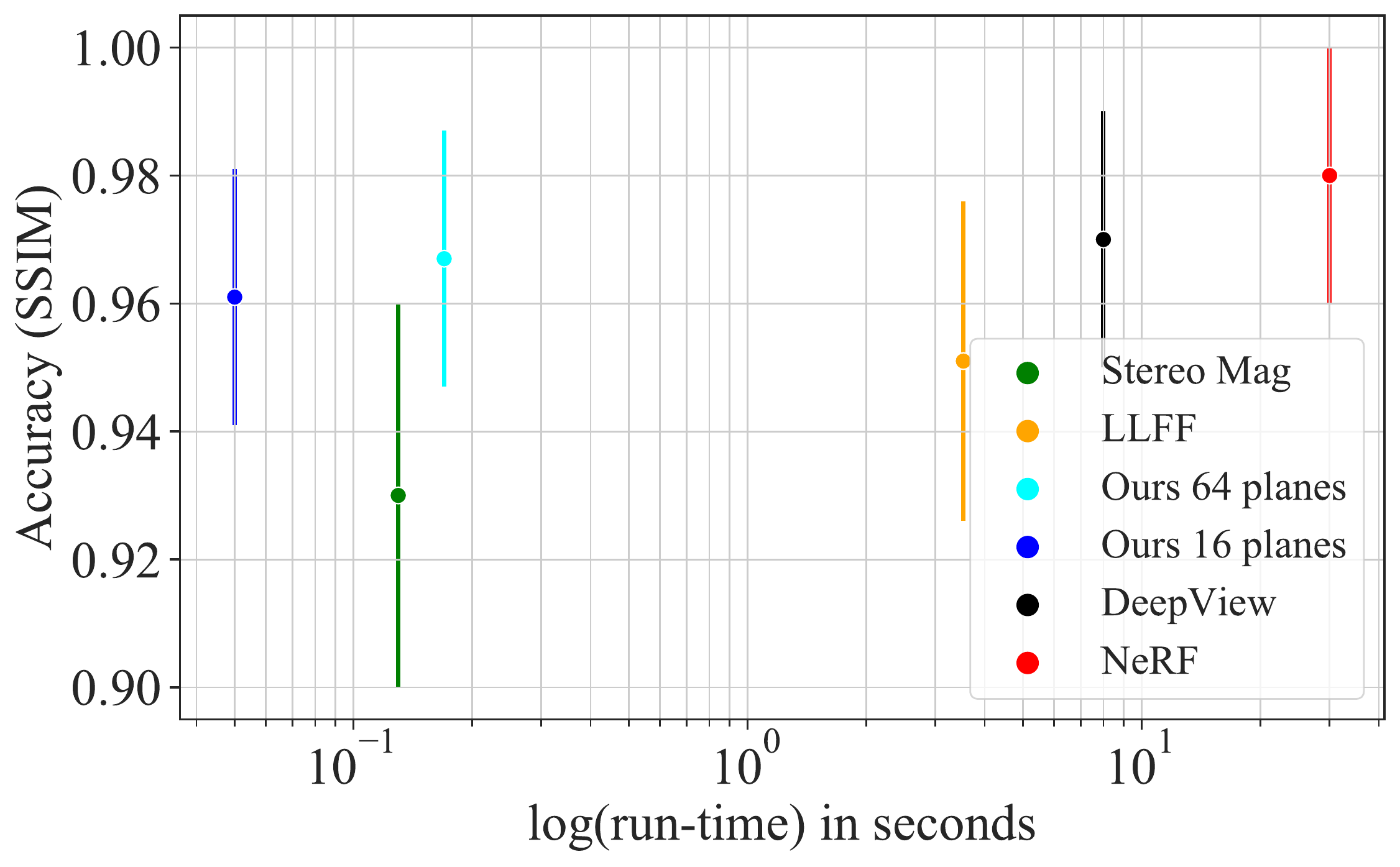}\par
    \caption{\small Accuracy vs run time for Stereo Mag ~\cite{zhou2018stereo}, LLFF~\cite{mildenhall2019local}, Ours 64 MPI planes, Ours 16 MPI planes, DeepView ~\cite{flynn2019deepview} and NeRF~\cite{mildenhall2020nerf}. SSIM accuracy is borrowed from ~\cite{mildenhall2020nerf} and ~\cite{flynn2019deepview}. Error bars indicate standard deviation across different tesing scenes. Run-time is in log scale for better interpretability. Our method with 16 plane MPI has the best run-time performance while achieving similar or better accuracy compared to other methods.}
    \label{fig:teaser_plot}
    \vspace{-1mm}
\end{figure}

Light fields offer a compelling way to capture the five-dimensional plenoptic function describing all possible light rays for a scene.
However, a dense light field captured at the Nyquist rate is practically challenging.
Recent work in view synthesis demonstrates a way to synthesize novel views from a sparse set of images. \cite{zhou2018stereo, mildenhall2019local, broxton:etal:siggraph20}.
These techniques enable applications in 360-degree imaging and immersive displays ~\cite{qian2018flare, sreedhar2016viewport} as well as real-time augmented and virtual reality (AR/VR) which can provide a live and interactive experience ~\cite{rolland1995comparison, rolland2000multifocal, chung1989exploring, janin1993calibration}. Live interactive viewing experiences require view synthesis algorithms with the following key attributes:
\vspace{-0.5em}
\begin{itemize} [noitemsep, leftmargin=*]
    \item Photo-realistic view synthesis, especially for nearby objects that produce large binocular disparity.
    \item Real-time processing of input views to generate target views at interactive frame rates.
\end{itemize}
\vspace{-0.5em}

Recently, Multi Plane Image (MPI) have become a favorable representation for real-time view synthesis of static scenes, achieving high quality results for many applications \cite{zhou2018stereo, srinivasan2019pushing, flynn2019deepview, tucker2020single}.
One key benefit of MPI representations is that they are computationally efficient to render, achieving real-time speeds for static scenes. However, state-of-the-art MPI methods require high-capacity neural networks and large amounts of computing power to generate the MPI from input views. Crucially, MPI generation is typically treated as a computationally expensive preprocessing step.

VR passthrough applications~\cite{chaurasia2020passthrough+} require view synthesis as the input cameras cannot be co-located with the viewer's eyes, and scene representations cannot be pre-processed. Consequently, there is a demand for a live view synthesis algorithm that can update MPIs as the scene changes over time. 
Due to the huge computational cost of generating an MPI, its use in such applications has been limited. 
Most current MPI generation methods have a fixed number of planes, where each plane has a fixed pixel disparity range, yet real-world scenes have variable disparity ranges that depend on scene depth and sensor resolution. To overcome this, current MPI-based methods require training with a high number of MPI planes and high capacity CNN to cover the large disparity range of real-world scenes.
This results in slower run-time performance.
To address all of these challenges, an entirely new method of generating an MPI is needed.
Specifically, a fast MPI generation technique must avoid disparity-dependent plane density and CNNs which require a large number of FLOPS.
A method that can dynamically generate MPI planes regardless of their number and spacing would allow for run-time efficiency by strategically placing fewer scene-dependent MPI planes.

In this paper, we propose a new MPI method with two novel properties, which we call \emph{dynamic target-centered} MPI.
First, we \emph{dynamically} generate the MPI one plane at a time at a depth chosen at run-time using multiple forward passes (or in batches), in contrast to existing work which generates all MPI planes, at preset depths, in one forward pass.
Second, we propose to generate the MPI directly at the target view (\emph{target-centered}), differing from a conventional MPI generated with respect to one input view. 
We show that these two changes allow us to use fewer parameters in the MPI generation network, in turn enabling fast run-time performance in spite of the need to evaluate the network for each generated MPI plane.

Our experiments demonstrate we can achieve state-of-the-art performance using an MPI representation for view synthesis with fewer floating-point operations per pixel (OPX) using a significantly smaller MPI generation network compared to existing MPI methods. 
We also show the dynamic selection of MPI planes can help us achieve high-quality results even with a low number of MPI planes at run-time, and can further boost the real-time performance.
In summary, we make the following technical contributions:

\begin{itemize} [noitemsep, leftmargin=*]
    \item We present a novel MPI method that can achieve state-of-the-art accuracy for MPI based view synthesis
    \item We show our MPI method allows for dynamic selection of the number and spacing between MPI planes at runtime (without re-training).
    \item To the best of our knowledge, we show the first MPI-based method, which can achieve photo-realistic live view synthesis and runs $70\times$ faster than Local Lightfield Fusion ~\cite{mildenhall2019local} (existing state-of-the-art using MPI) with the same or better pixel accuracy.
\end{itemize}
\section{Related Work}
\label{sec:related}

View synthesis is a well-studied problem in computer vision and computer graphics.
Early attempts at view synthesis involve simple pixel interpolation~\cite{fleten1993shenchang} of images taken from different views.
Levoy et al.~\cite{levoy1996light} developed an image-based rendering technique by sampling the plenoptic function using multiple views.
While these methods work well for low-resolution images, good quality results depend on the availability of dense views as well as small disparities. 

\paragraph{Image Based Rendering} IBR ~\cite{shum2000review} is a broad class of view synthesis methods comprising of techniques ranging from interpolation of lightfields and the plenoptic function ~\cite{adelson1991plenoptic, levoy1996light} to an explicit estimation of the 3D shape, appearance, and scene geometry ~\cite{zitnick2004high, vedula2005three, shan2013visual}.
Lightfield interpolation methods are simple and inexpensive, but require dense sampling of the input views, while explicit scene geometry estimation is complex and computationally expensive, but works with fewer input views.
A typical pipeline involves depth estimation from multi-view input and subsequent warping to achieve view synthesis~\cite{chaurasia2013depth,chaurasia2011silhouette}.
However, these methods are only as good as the underlying depth estimation and forward projection warping method used.
More recently, machine learning techniques, particularly those employing deep learning, are able to train on large datasets of image pairs to learn the synthesis of novel views.
One such method ~\cite{kalantari2016learning} uses two separate networks, one for disparity estimation and another for fusing the disparity warped input views to target view.
However, techniques dependent on backward projection warping are poor at handling disoccluded regions, especially with few input views.

\paragraph{Implicit Representations} 
Recently there is remarkable progress using Multi-Layer Perceptrons (MLPs) to represent the scene as a continuous function, e.g. NeRF \cite{mildenhall2020nerf} and its variants \cite{liu:etal:neuralips20}. Recent methods extend NeRF to leverage internet images \cite{martin:etal:arXiv20} and freely captured images \cite{zhang:etal:arXiv2020}.
Such representations have demonstrated state-of-the-art results for multi-view synthesis, with challenging geometry, lighting, and view-dependent effects. However, these approaches require tens of GPU computation hours to train one MLP per scene. Rendering a high-resolution image with a pre-trained MLP model can take minutes. Although recent work \cite{liu:etal:neuralips20} improved the rendering speed with local volumetric priors, the speeds achieved are still not suitable for any real-time applications.

\paragraph{Multi Plane Image (MPI)} MPI representation offers the best combination of speed and quality.
The scene is represented using a number of planar images.
Each plane is associated with a depth/disparity and the image is represented by four values (3 color and 1 alpha) at each pixel.
Novel view synthesis from an MPI is performed by simply warping the depth planes to a novel view (via homography) and back-to-front alpha compositing to generate the target view.
Since this involves only a few matrix operations, rendering from an MPI is very fast, especially on a GPU.
Zhou \etal ~\cite{zhou2018stereo} used a CNN-based approach to generate an MPI from input views in a single forward pass. Since then a number of MPI-based approaches ~\cite{srinivasan2019pushing, mildenhall2019local} have been proposed and shown to reconstruct high-fidelity novel views in a single shot.
Flynn \etal \cite{flynn2019deepview} generate MPIs for a scene by iteratively refining the MPI to fit input views using simple gradient-based optimization methods or learned gradients for fast convergence.

Our method is essentially an MPI-based method. However, our MPI generation differs from existing approaches by rendering directly at the target view and generating RGBA values for each MPI plane individually (or batch processing for efficiency). 

\section{Preliminary Background}
\label{sec:multi_plane_image}
A Multi-Plane Image (MPI) represents a scene using a set of $D$ depth planes from a camera viewpoint $v_{mpi}$.
Each MPI plane consists of four channels (three RGB and an alpha channel)
Conventional MPI methods input $V$ RGB images and output $D$ RGBA images. They first generate a plane sweep volume (PSV) with $D$ depth planes from the input images.
The PSV is formed by applying homography warping from the input camera view to the target MPI camera view for a few pre-determined depths for all input images.
Before passing the input to a CNN, these warped images are either concatenated along the channel dimension ~\cite{zhou2018stereo} for 2D CNN or stacked along a separate depth dimension for input to a 3D CNN ~\cite{srinivasan2019pushing, mildenhall2019local}.
Existing MPI generation methods rely on two significant assumptions:
\begin{itemize} [noitemsep, leftmargin=*]
\item \textbf{Static MPI:} Early MPI methods \cite{zhou2018stereo} assumed the same number and spacing of planes at training and rendering. Recent 3D CNN approaches ~\cite{mildenhall2019local} allow for varying numbers of uniformly spaced MPI planes, but the performance quickly degrades when the number and spacing of planes differ in training and rendering. We perform an ablation study in \secref{sec:ablation_study} and \figref{fig:static_dynamic_variable_planes} summarizes this finding.
\item \textbf{Input centered MPI:} Existing MPI methods assume that the MPI planes will only be generated once with an input reference view (pre-generation). Then a novel view can be quickly rendered to various target views at run-time via back-to-front composition.
\end{itemize}
\vspace{-0.5em}

However, the assumption that MPIs can be precomputed to increase run-time efficiency only holds for pre-recorded scenes intended for offline playback.
For live view synthesis, each input-centered MPI is used only once to generate the target novel view(s) at each time instance.
In this case, the total time to render a novel view for a dynamic scene is the sum of MPI generation and novel view render time. For existing methods, this time is dominated by the exorbitant computational cost of MPI generation.

\section{Proposed Method}
\label{sec:method}

\begin{figure}
    \centering
    \includegraphics[width=0.45\textwidth, keepaspectratio]{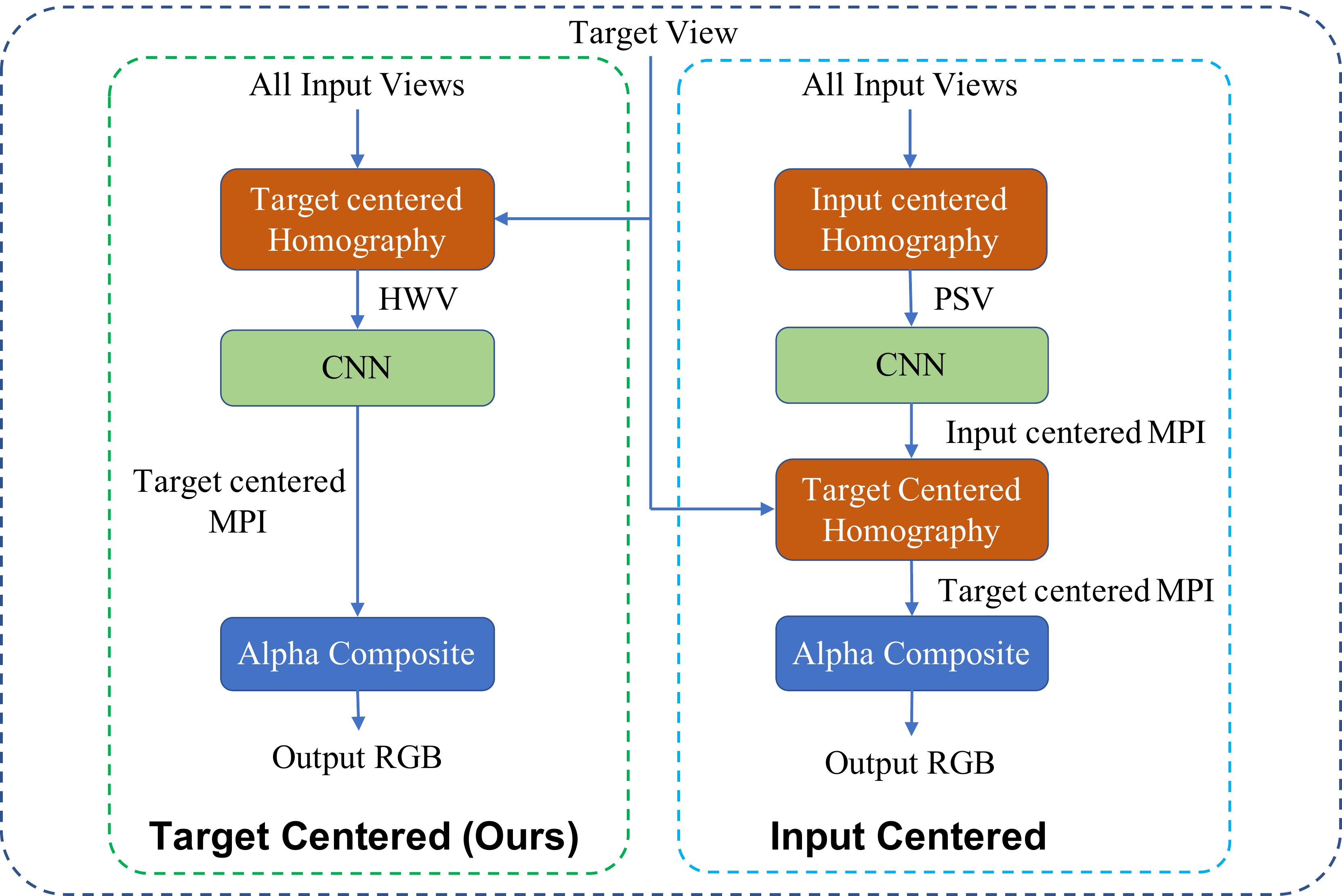}\par
    \caption{\small \textbf{A block schematic of our approach}. Our target-centered MPI generation (left) warp the input images to the target view via homography prior to passing to the network (a homography warped volume, or HWV), whereas conventional input-centered MPI generation (right) transforms the input views to a reference input view, which is passed to the network (a plane sweep volume, or PSV), and then applies a homography to transform to the output view.}
    \label{fig:comparison_diag}
    \vspace{-1em}
\end{figure}

We propose a novel MPI method that can address the limitations in conventional MPI representations to enable the fast run-time generation and maintain high reconstruction fidelity for live view synthesis.
Our method differs from existing MPI generation and subsequent view synthesis in two distinct ways (see Fig. \ref{fig:comparison_diag}):
\vspace{-0.5em}
\begin{itemize} [noitemsep, leftmargin=*]
    \item \textbf{Dynamic MPI}: In contrast to static MPI, which uses a CNN to output all RGBA images (or the corresponding blending weights) at all planes in a single forward pass, our network generates a single RGBA image in a single run.
    We run the network multiple times to generate all the planes of the MPI.
    We can dynamically choose the number and spacing of the MPI planes at run-time, which existing methods cannot. 
    \item \textbf{Target-centered MPI}: In contrast to input-centered MPI, which first generates MPI at a fixed input camera pose and then warps the MPI planes to the target view at run-time, we first warp input images to a target view (target centered), and then pass it through a CNN. 
\end{itemize}
\vspace{-0.5em}

Target-centered MPI generation results in better reconstruction quality, even with a relatively shallow CNN, allowing for fast evaluation.
Dynamic MPI generation allows us to learn a depth plane agnostic network that generalizes well when the MPI planes (number and spacing) do not match during input and output.
Dynamic MPI generation also enables dynamic plane selection, which further improves run-time performance without sacrificing accuracy.

\begin{figure*}[!htb]
    \centering
    \includegraphics[width=0.95\textwidth, keepaspectratio]{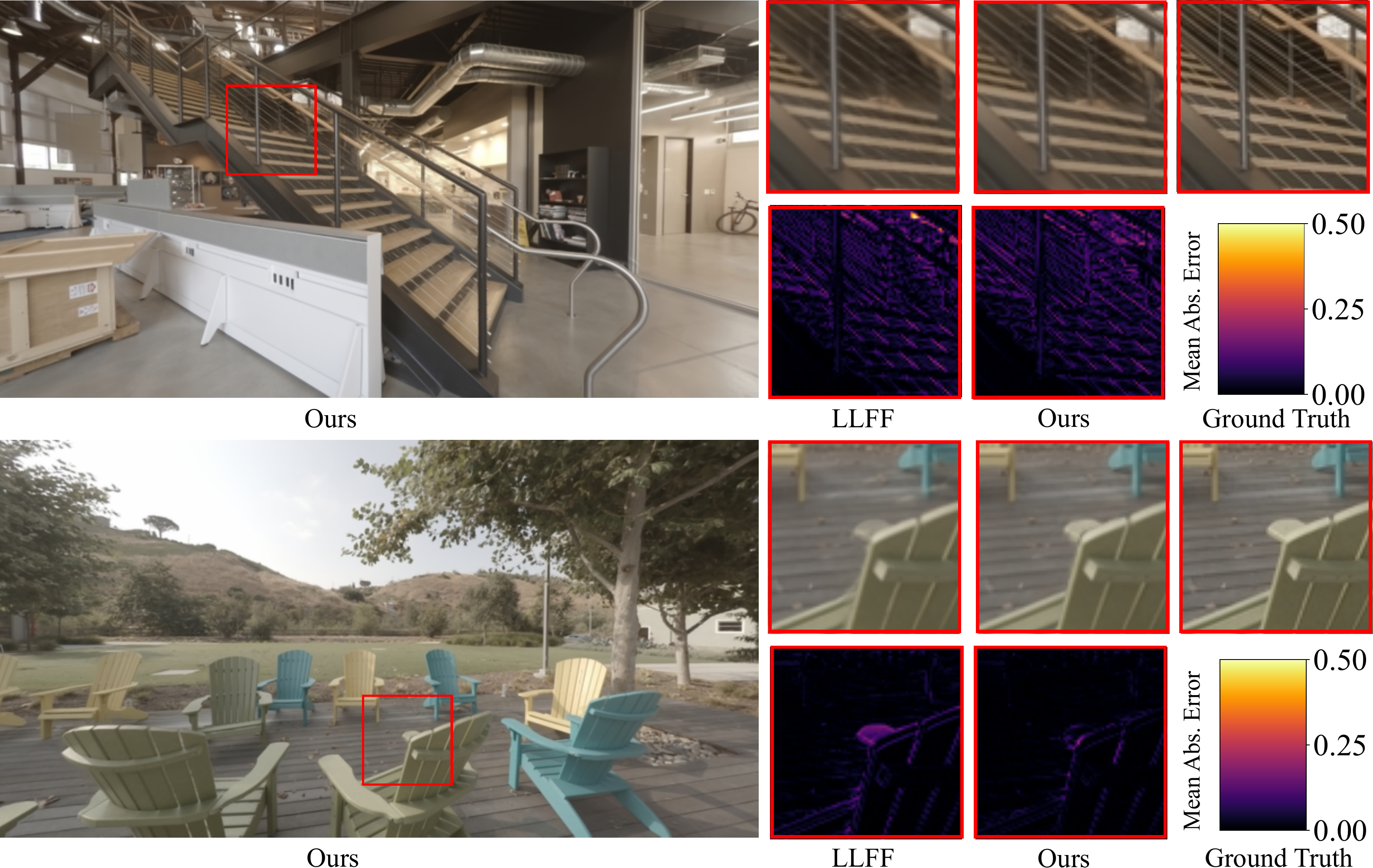}\par
    \caption{\small Reconstruction results for 10-view Spaces dataset test scenes. Insets show a comparison against Local Lightfield Fusion (LLFF) \cite{mildenhall2019local} using two 5-view MPI and absolute difference from ground truth. Note our method generates better results around occlusion boundaries in the first scene and thin structures in the second scene. Supplementary materials contain video results for continuous viewpoint motion.}
    \label{fig:comparison_llff}
    \vspace{-1.5em}
\end{figure*}

\paragraph{Overall Pipeline} To generate a target view with $D$ MPI planes, we run $D$ forward passes, one for each of the $D$ MPI planes.
In each forward pass, we first warp all input images to the target view to generate a homography warped volume (HWV).
We pass HWV volume to a CNN to generate the blending weights and alpha corresponding to a single depth plane.
We then perform a weighted alpha composite of the blending weights, alpha, and HWV to render the target image.
We describe the key steps in the following.

\paragraph{Homography Warped Volume (HWV)} 
First, we select $D$ disparity (eqi-disparity for simplicity) planes.
The input views are warped to target view for each disparity plane using homography warping, resulting in an HWV $\mathbf{\Phi} \in \mathbf{R}^{D \times V \times 3 \times H \times W}$, where $H$ and $W$ are the height and width of the input images.
This aligns pixels at a specific depth when viewed from the target viewpoint while misaligning the pixels from different depths.
The following network is only expected to learn to assign high alpha values (and RGB weights) to pixels that align well, and low alpha values to pixels that don't align well.
This is a relatively simple mapping to learn compared to generating output views from the plane sweep volume that is conventionally used.

\paragraph{Neural Network} 
We use a 2D CNN $f_\theta$ to learn the weights $\mathbf{W} \in \mathbf{R}^{D \times (V-1) \times H \times W} $ and alpha $\alpha \in \mathbf{R}^{D \times H \times W}$ corresponding to the target MPI from the input HWV $\Phi$
\begin{align}
    f_\theta: \Phi \rightarrow \mathbf{W}, \alpha
\end{align}

We use a standard U-Net structure with five convolution layers followed by a transpose convolution with a skip connection. Given the network output tensor, we use a sigmoid operation on the first channel to predict $\alpha$ and a softmax layer on the other to generate the weights $\mathbf{W}$.
We show the network architecture in Tab. \ref{table:network_architecture}. 

\begin{table}[t]
    \begin{center}
        \begin{tabular}{ccccccc}
            \hline
            \textbf{Layer} & \textbf{k} & \textbf{s} & \textbf{chns} & \textbf{in} & \textbf{out} & \textbf{input} \\
            \hline
            conv1 & 3 & 1 & 3$V$ / 16 & 1 & 1 & PSV \\
            conv2 & 3 & 2 & 16 / 32 & 1 & 2 & conv1 \\
            conv3 & 3 & 2 & 32 / 64 & 2 & 4 & conv2 \\
            conv4 & 3 & 2 & 64 / 128 & 4 & 8 & conv3 \\
            conv5 & 3 & 1 & 128 / 128 & 8 & 8 & conv4 \\
            \hline
            ups5 &   &   & 128 / 128 & 8 & 4 & conv5 \\
            conv6 & 3 & 1 & 192 / 64 & 4 & 4 & ups5 + conv3 \\
            ups6 &   &   & 64 / 64 & 4 & 2 & conv6 \\
            conv7 & 3 & 1 & 96 / 32 & 2 & 2 & ups6 + conv2 \\
            ups7 &   &   & 32 / 32 & 2 & 1 & conv7 \\
            conv8 & 3 & 1 & 48 / 16 & 1 & 1 & ups7 + conv1 \\
            conv9 & 3 & 1 & 16 / $V$ & 1 & 1 & conv8 \\
            \hline
        \end{tabular}
        \vspace{3mm}
        \caption{\small \textbf{Network architecture}. \textbf{k} is the kernel size, \textbf{s} is the stride, \textbf{chns} denote the input and the output channels for the corresponding layer. We also show \textbf{in} and \textbf{out} denoting the accumulated output stride of each layer. Layers named \textit{ConvX} are convolutional layers followed by batchnormalization and ReLU activation (except last layer which is just a convolutional layer), whereas layers named \textit{upsX} are bilinear upsampling layers in the spatial dimension.
        The input has 3$V$ channels and output has $V$ channels, where $V$ is the number of input views.}
        \label{table:network_architecture}
        \vspace{-5mm}
    \end{center}
\end{table}

\paragraph{Rendering the MPI} Given predicted $\mathbf{W}, \alpha$, we first normalize $\mathbf{W}_i$ based on the Euclidean distance of the input view the target, and then blend using alpha composition to generate the final RGB image at the target view.





\paragraph{Dynamic Plane Selection} Our method allows for preferentially choosing depth planes at run-time.
Ideally, we want to have MPI planes at depths that match the object's position in the scene.
We employ a simple depth histogram-based strategy to select $K$ ($< D$) planes from $D$ uniform disparity MPI planes.
Each output pixel is assigned a depth index $d_i$ based on the depth plane its alpha value achieves maximum value.
We then create a histogram of the depth index for all pixels by binning them into $D$ bins.
Depth planes corresponding to top $K$ bins are selected for rendering a novel view.
We plot the performance of the view synthesized with these $K$ MPI planes.
\figref{fig:plane_selection} shows that the performance of dynamically selected planes is more favorable than uniform disparity planes for the same number of MPI planes.

We demonstrate this plane selection scheme can be particularly useful for live video synthesis in \secref{sec:video_view_synthesis}, where we use the first frame of a video to determine the depth planes to use for the rest of the sequence.





\section{Experiments}
\label{sec:experiments}

\begin{table*}[th]
    \begin{center}
        \begin{tabular}{|c|l|ccc|ccc|}
            \hline
            \multicolumn{2}{|c|}{\textbf{Model}} & \multicolumn{3}{c|}{\textbf{Deepfocus}} & \multicolumn{3}{c|}{\textbf{Spaces}} \\
            \hline
            & & PSNR $\uparrow$ & SSIM $\uparrow$ & LPIPS $\downarrow$ & PSNR $\uparrow$ & SSIM $\uparrow$ & LPIPS $\downarrow$ \\
            \hline
            \multirow{3}{*}{\textbf{4-View}} & Soft3D\cite{penner2017soft} & - & - & - & 30.14 & 0.9273 & 0.1177 \\
            & DeepView ~\cite{flynn2019deepview} (80 planes) & - & - & - & \textbf{32.92} & \textbf{0.9607} & \textbf{0.0766} \\
            & Ours & 31.98 & 0.9576 & 0.0743 & 32.01 & 0.9475 & 0.0992 \\
            \hline
            \multirow{3}{*}{\textbf{5-View}} & Stereo Magnification\cite{flynn2016deepstereo} & 29.23 & 0.9301 & 0.0769 & 27.98 & 0.8928 & 0.1231 \\
            & Local LF \cite{mildenhall2019local} & 33.8 & 0.9611 & 0.0581 & 30.44 & 0.9324 & 0.1181 \\
            & \textbf{Ours} & \textbf{34.29} & \textbf{0.9678} & \textbf{0.0565} & \textbf{31.05} & \textbf{0.947} & \textbf{0.1017} \\
            \hline
            \multirow{2}{*}{\textbf{10-View}} & LLFF\cite{mildenhall2019local} & 36.08 & 0.9841 & 0.0421 & 33.06 & 0.949 & 0.1149  \\
            & \textbf{Ours} & \textbf{36.69} & \textbf{0.9863} & \textbf{0.0402} & \textbf{34.18} & \textbf{0.9631} & \textbf{0.085} \\
            \hline
        \end{tabular}
        \vspace{3mm}
        \caption{\small Performance summary for different view synthesis methods on the Deepfocus~\cite{deepfocus_dataset} and Spaces~\cite{flynn2019deepview} datasets. Results shown for 4, 5, and 10 input views to provide a fair comparison. 5 views are used for Stereo Magnification~\cite{zhou2018stereo} and Local LF (base network for LLFF ~\cite{mildenhall2019local}), which achieve best results when the MPI is centered at one of the input views. For LLFF, we need 2 MPIs generated using Local LF, requiring 10 views. The number of MPI planes was fixed at 64 eqi-disparity planes, except for DeepView, which had 80 MPI planes. Our method outperforms Soft3D, Stereo Magnification, Local LF, LLFF, while achieving comparable results to DeepView, an iterative method.}
        \label{table:comparison}
    \end{center}
    \vspace{-3mm}
\end{table*}

\paragraph{Implementation} Our method involves homography warping input views to target view, followed by forward pass of our CNN described in table \ref{table:network_architecture}for each depth plane and subsequently weighted alpha compositing.
For training our CNN network, we use Adam optimizer with a learning rate of $10^{-4}$, $\beta_1=0.9$ and $\beta_2=0.999$.
For results on the Deepfocus dataset, each network is trained on the Deepfocus dataset with the specific input view configuration for 100K iterations.
For results on Spaces dataset, network pre-trained on Deepfocus dataset were used as initialization and trained on Spaces dataset for another 100K iterations. 

\paragraph{Baselines} We compare to the following methods: 
\begin{itemize} [noitemsep, leftmargin=*]
\item Stereo Magnification~\cite{zhou2018stereo}: We modified it to take 5 input views.
\item Soft3D~\cite{penner2017soft}: This method provides a baseline using volume rendering without learning. We use the test images generated by authors of Soft3D on Spaces dataset with 4 views and a large baseline ~\cite{flynn2019deepview}.
\item Local Light field Fusion (LLFF) ~\cite{mildenhall2019local} : This method takes multiple neighboring pre-rendered MPIs to synthesize a specific view at run-time. 
\item Local Lightfield (Local LF): It is the 3D CNN used to generate MPIs at specific grid points for LLFF \cite{mildenhall2019local}.
\item DeepView~\cite{flynn2019deepview}: It provides a baseline for using iterative learned gradient descent to generate MPI planes. We use the test images generated by authors on Spaces dataset with 4-view, large baseline, and 80 MPI planes ~\cite{flynn2019deepview}.
\end{itemize}

We trained and tested our implementation of Stereo Magnification, LocalLF, and LLFF methods on Spaces and Deepfocus datasets.
We couldn't compare Soft3D and DeepView with other camera configurations and Deepfocus datasets as there is no existing public implementation of their method. 

Our method is flexible with the number of input views.
We perform experiments with three different input camera configurations with a different number of input views. 
We evaluate on 4 input views to match the 4-view, large baseline configuration used in DeepView ~\cite{flynn2019deepview} and compare against DeepView and their implementation of Soft3D on Spaces dataset.
We use five input views to compare performance against modified Stereo Magnification ~\cite{zhou2018stereo} and Local Lightfield 3D CNN ~\cite{mildenhall2019local} which requires the MPI to be centered at one input view for best performance.
We use ten input views to compare against Lightfield Fusion ~\cite{mildenhall2019local} which requires at least two MPIs as input, each requiring five input views.


\paragraph{Metrics} We provide a quantitative evaluation in Tab. \ref{table:comparison} using standard image metrics PSNR and SSIM as well as perceptual metric LPIPS ~\cite{perceptualsimilarity2018}.

\paragraph{Datasets} We test our performance on two datasets: 
\begin{itemize} [noitemsep, leftmargin=*]
    \item \textbf{Deepfocus\cite{deepfocus_dataset}:} It consists of 86 synthetic rendered scenes of objects with challenging texture and lighting conditions. We perform all ablations on this dataset.
    \item \textbf{Spaces\cite{flynn2019deepview}: } It consists of 100 real world (outdoors + indoors) from camera rig of 16 cameras.
\end{itemize}

We first provide both quantitative and qualitative evaluation of our method compared to baselines on existing static novel view synthesis datasets in \secref{sec:novel_view_synthesis}, and then we further demonstrate a few ablations in \secref{sec:ablation_study}. We evaluate our method for speed by comparing the run-time performance in \secref{sec:speed}.
We also show the performance of our video view synthesis in \secref{sec:video_view_synthesis} using the rendered scene from the First Contact used in Deepfocus \cite{deepfocus_dataset}. We include the video results in the supplementary materials.


\begin{figure*}[!htb]
    \centering
    \includegraphics[width=0.95\textwidth, keepaspectratio]{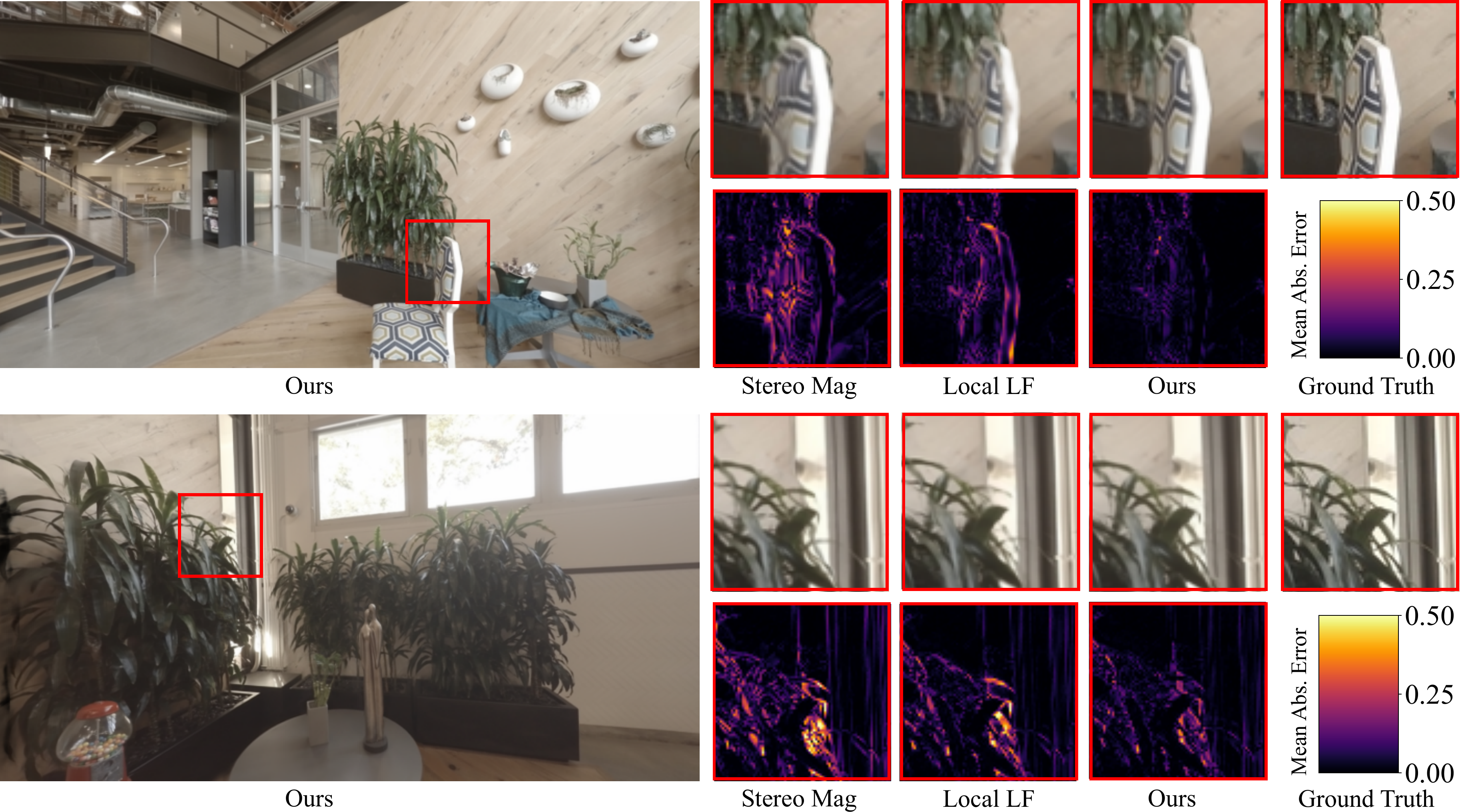}\par
    \caption{\small Reconstruction results of our method on a 5-view test scene from the Spaces dataset. Insets show a comparison against Stereo-Magnification \cite{zhou2018stereo} and Local Lightfield (without fusion) using 5-view MPIs \cite{mildenhall2019local}, as well as absolute difference from ground truth. Supplementary materials contain video results for continuous viewpoint motion.}
    \label{fig:comparison_5_view}
\end{figure*}
\begin{figure*}[h]
    \centering
    \includegraphics[width=\textwidth, keepaspectratio]{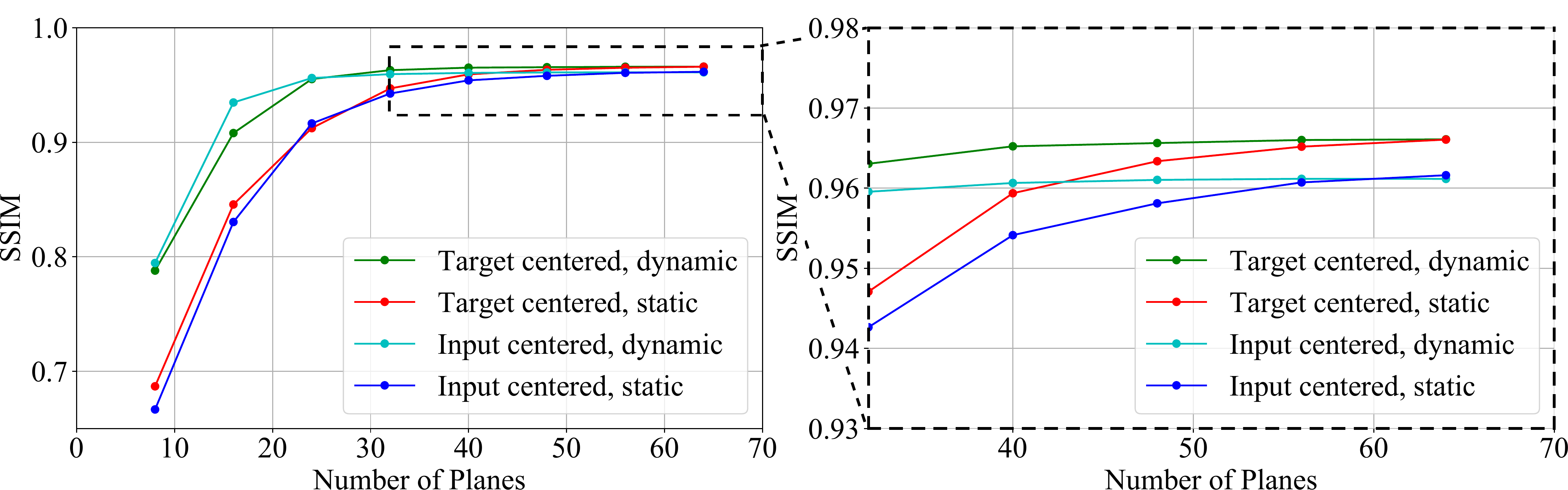}
    \caption{\small Performance of all four combination of input-centered / target-centered and static / dynamic MPI generation methods.
    All method were trained with 64 MPI planes and tested with variable number of planes ranging from 8 to 64.
    We find that target-centered approach helps achieve better peak accuracy, while dynamic MPI generation shows a favourable falloff when training and testing number of planes do not match.}
    \label{fig:target_input_static_dynamic}
\end{figure*}

\subsection{Evaluation on Novel View Synthesis}
\label{sec:novel_view_synthesis}

\tabref{table:comparison} shows a summary of our performance of ours as well as other state-of-the-art methods on different datasets. It shows the superiority of our reconstructions in terms of PSNR, SSIM, and LPIPS metrics. Continuous view interpolation video results are included in the supplementary materials.

We show visual comparisons of the 10 view input with LLFF in \figref{fig:comparison_llff} as well as a 5 view input comparison with Local LF and Stereo-Magnification in \figref{fig:comparison_5_view}. Zoomed in insets with the corresponding absolute error map shows that our method achieves better results in challenging regions like occlusion boundaries and thin structures. 

\subsection{Ablation Study}
\label{sec:ablation_study}

We perform a few ablations to better understand our \emph{dynamic target-centered} MPI.
We use the same network in \tabref{table:network_architecture} as comparing it against the 3D CNN architecture of Local Lightfield Fusion ~\cite{mildenhall2019local} which we refer to as Local LF.

\paragraph{Combinations of MPI representation.} 
Fig. \ref{fig:target_input_static_dynamic} shows the performance of all four combinations of target-centered / input-centered and static/dynamic plane generation.
The zoom inset on the right shows that target-centered methods improve accuracy, while the trend of static vs dynamic plots on the left shows that dynamic processing improves performance when the training and testing number of planes are different.

\begin{figure}[h]
    \centering
    \includegraphics[width=0.45\textwidth, keepaspectratio]{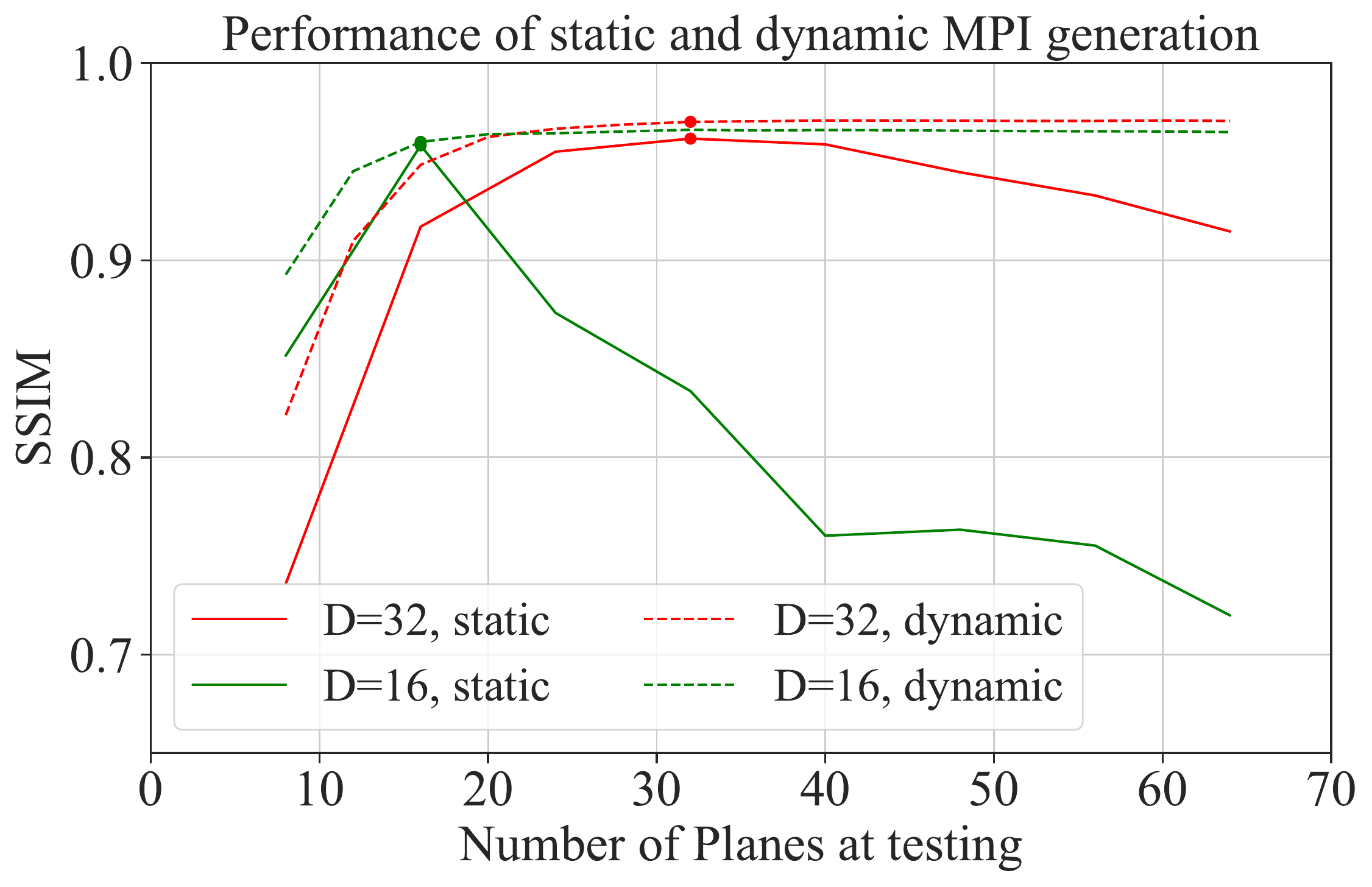}
    \caption{\small Performance of target-centered static (Local Lightfield 3D CNN adapted for target-centered MPI generation) vs our target-centered dynamic generation by training networks with different number of equi-disparity planes. Solid colored dot indicates where training and testing plane counts match. Note that dynamic MPI generation shows a more favorable falloff when the number of planes decreases from 64 to 8 uniformly spaced disparity planes.}
    \label{fig:static_dynamic_variable_planes}
\end{figure}

\paragraph{Varying number of output planes} Figure \ref{fig:static_dynamic_variable_planes} shows the performance of static vs dynamic networks on different numbers of output planes when trained for 32 planes (red curve) and 16 planes (green curve).
We see that the performance of Local LF peaks when tested on the same number of planes as training but quickly degrades with a change in the number of testing planes. Our dynamic MPI generation method yields a monotonically improving performance as we increase the number of testing planes.

\subsection{Speed}
\label{sec:speed}

The primary advantage of our MPI method is its fast run-time performance compared to other view synthesis methods when considering the time taken to generate the MPI.
Tab. \ref{table:speed} shows the performance of modified Stereo Magnification, Local LF (base network for LLFF), LLFF, and our method on 64, 32, and 16 planes, on an image resolution of 350 x 500.
We benchmark it via a Pytorch implementation using 32-bit floating-point precision on a single Nvidia Quadro V100 GPU.
Tab. \ref{table:speed} shows that our method is significantly faster compared to all existing MPI methods, requiring significantly fewer parameters to learn, and significantly fewer OPX at run-time.

\begin{figure}[h]
    \centering
    \includegraphics[width=0.45\textwidth, keepaspectratio]{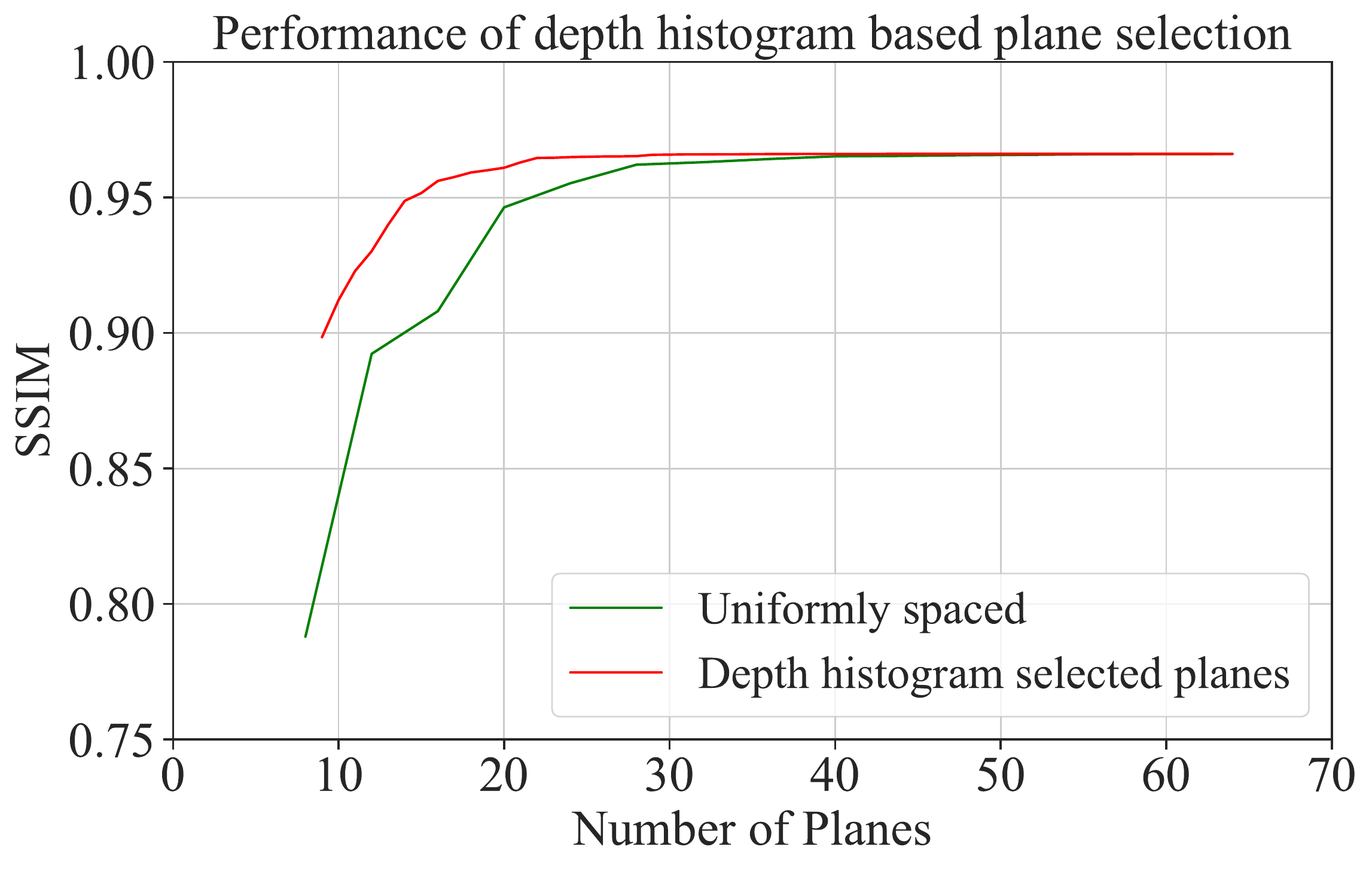}\par
    \caption{\small Performance comparison of our depth histogram based dynamic plane selection method described in \secref{sec:speed} vs uniformly spaced disparity planes using the same network.
    The network was trained using 64 uniform disparity planes.
    We see that dynamic plane selection improves performance with fewer planes while matching performance with many planes.}
    \label{fig:plane_selection}
    \vspace{-3mm}
\end{figure}

Fig. \ref{fig:plane_selection} shows the importance of dynamic plane selection discussed in Sec. \ref{sec:method}.
By using our dynamic plane rendering, we achieve an SSIM score of 0.9575 with only $K$ = 16 planes, which is less than a 1\% drop from the performance of 0.966 at 64 uniform disparity planes (same as training).
This essentially allows us to render a static scene in about 50 ms, corresponding to 20 FPS.
Compared to Local Lightfield Fusion, this is about 70 times faster while achieving higher accuracy.

\begin{table}
    \begin{center}
    \resizebox{\columnwidth}{!}{%
        \begin{tabular}{|l|c|c|c|}
            \hline
            Model & run-time (s) $\downarrow$ & OPX $\downarrow$ & Parameters \\
            \hline
            Stereo Magnification (64 planes) & 0.127 & 1.39 M & 14.9 M \\
            Local LF \cite{mildenhall2019local}(64 planes) & 1.73 & 2.96 M & 835 K \\
            LLFF \cite{mildenhall2019local} (64 planes) & 3.53 & 5.92 M & 835 K \\
            Ours (64 planes) & 0.172 & 1.79 M & 391 K \\
            Ours (32 planes) & 0.095 & 894 K & 391 K \\
            Ours (16 planes) & \textbf{0.050} & 447 K & 391 K \\
            \hline
        \end{tabular}
        }%
        \caption{\small Performance without pre-generating the MPI, in terms of run time, OPX (FLOPS per pixel), and number of parameters. Our approach has significantly lower complexity and better run-time speed.}
        \label{table:speed}
    \end{center}
    \vspace{-5mm}
\end{table}

\subsection{Video View Synthesis}
\label{sec:video_view_synthesis}

We demonstrate novel view synthesis on videos using the First Contact robot scene from the Deepfocus dataset \cite{deepfocus_dataset} in our supplementary materials. Our method can achieve about 20FPS rendering with dynamic plane selection with minimal loss in accuracy. 
Optimally selecting MPI planes requires prior knowledge about a scene (e.g., from a depth sensor or densely sampling the scene).
For a proof-of-concept with video input, we sample 64 uniform disparity planes for the first frame, then downselect to 16 planes using our dynamic plane selection method described in  \secref{sec:method} and then reuse them for the remainder of the video.
This amortizes the computational cost of generating a large number of planes for the first frame.
Our experiments show less than 1\% loss in SSIM scores using this dynamic plane strategy compared to running all 64 MPI planes frame-by-frame.


\section{Limitations}
\label{limitations}

Though our method is favorable for live view synthesis where pre-generating MPI is not feasible, for applications including static scene rendering or video replay, rendering can still be more efficient using existing MPI methods or preprocessing MPI videos to a more compact representation\cite{broxton:etal:siggraph20}, compared to our method. 
Our method as well as existing methods also require the orientation of the input cameras to be fixed at training and testing and hence requires re-training for different camera configurations.

\section{Conclusion and Future Work}
\label{conclusion}
We propose a novel MPI representation to achieve high-quality view synthesis compared to existing techniques that use a pre-generated MPI.
We show that our method achieves a better quality of rendering results compared to existing state-of-the-art methods on both synthetic as well as real-world datasets while being orders of magnitude faster.
We further show that our method offers the flexibility to choose MPI depth planes at run-time without significant performance degradation.
By combining our fast view synthesis with a histogram-based depth selection technique, we were able to achieve real-time video view synthesis.

We believe that sampling a scene at run time with dynamically selected MPI planes can further boost performance for real-time live view synthesis applications. 
Lastly, our results point to a promising direction for future improvements exploiting spatio-temporal coherence within MPI representations, thereby reducing complexity in generating separate MPIs for successive frames. 
We leave this for future work.

{\small
\bibliographystyle{ieee_fullname}
\bibliography{main}
}

\end{document}